\documentclass{article} 
\usepackage{iclr2026_conference,times}

\usepackage{amsmath,amsfonts,bm}

\def\eqref#1{equation~\ref{#1}}

\def\1{\bm{1}}

\DeclareMathAlphabet{\mathsfit}{\encodingdefault}{\sfdefault}{m}{sl}
\SetMathAlphabet{\mathsfit}{bold}{\encodingdefault}{\sfdefault}{bx}{n}

\usepackage{graphicx}
\usepackage{booktabs}
\usepackage{multirow}
\usepackage{hyperref}
\usepackage{url}
\usepackage{framed}
\usepackage{tcolorbox}
\usepackage{svg}
\usepackage{algorithm}
\usepackage{algpseudocode}
\usepackage{amsmath}
\usepackage{enumitem}
\usepackage{amssymb}
\usepackage{multirow}
\usepackage{tcolorbox}
\tcbuselibrary{breakable, listingsutf8}  
\usepackage{listings}
\usepackage{adjustbox}
\usepackage{listings}
\usepackage{xcolor}

\usepackage{caption}  

\title{MTIR-SQL: Multi-turn Tool-Integrated  Reasoning Reinforcement Learning for Text-to-SQL}
\lstdefinestyle{json}{
  language=JSON,
  basicstyle=\ttfamily\small,
  showstringspaces=false,
  columns=flexible,
  keepspaces=true
}
\author{Zekun Xu$^{1}$\thanks{Equal contribution.}, Siyu Xia$^{2,3}$\footnotemark[1], Chuhuai Yue$^{4}$, Jiajun Chai$^{1}$\thanks{Project Leader}, Mingxue Tian$^{5}$, Xiaohan Wang$^{1}$,\\ \textbf{Wei Lin}$^{1}$,  \textbf{Haoxuan Li}$^{6}$,  \textbf{Guojun Yin$^{1}$}\thanks{Corresponding author.Contact: xuzekun02@meituan.com, yinguojun02@meituan.com}\\
  $^{1}$Meituan \hspace{5pt}  $^{2}$Institute of Automation, Chinese Academy of Sciences \\
  $^{3}$School of Artificial Intelligence, University of Chinese Academy of Sciences\\
  $^{4}$Beijing Institute of Technology \hspace{2pt}
  $^{5}$Shanghai Jiao Tong University \hspace{2pt}
  $^{6}$Peking University
}
\iclrfinalcopy 

\begin{document}

\maketitle

\begin{abstract}
As large language models (LLMs) are increasingly used in Text-to-SQL tasks, Reinforcement Learning (RL) has become a common method for improving performance. Existing methods primarily rely on static execution feedback, which restricts real-time error correction. However, integrating multi-turn tool invocation along with dynamic feedback could significantly improve adaptability and robustness, ultimately enhancing model performance. To address these issues, we propose \textbf{MTIR-SQL, an innovative Multi-turn Tool-Integrated Reasoning reinforcement learning framework for Text-to-SQL}. Our approach introduces an execution-aware multi-turn reasoning paradigm that seamlessly incorporates database execution feedback at each reasoning step, enabling context-sensitive query generation and progressive refinement throughout the reasoning process. The framework extends the GRPO algorithm to accommodate complex multi-turn interaction scenarios. Considering the training instability characteristics of MTIR and the potential for significant Deviation of model distribution from the initial model, we enhance the GRPO algorithm by adding a trajectory filtering mechanism and removing KL loss constraints. Experimental results demonstrate that MTIR-SQL, with 4B parameters, achieves \textbf{64.4}\% accuracy in the BIRD Dev and \textbf{84.6}\% execution accuracy in the SPIDER Dev, significantly outperforming existing approaches. 

\end{abstract}

\section{Introduction}
Text-to-SQL, the task of automatically translating natural language questions into executable SQL queries, is a key technique for lowering the barrier to database access~\citep{liu2025surveytexttosqlerallms}. By enabling non-technical users to query structured data in natural language, it has found wide applications in business intelligence, data analytics, and interactive question answering~\citep{hong2025nextgenerationdatabaseinterfacessurvey}.

Existing approaches to Text-to-SQL generally fall into three paradigms: (i) Supervised Fine-Tuning (SFT) of domain-specific open-source models~\citep{li2025omnisqlsynthesizinghighqualitytexttosql,li2024codesbuildingopensourcelanguage}; (ii) prompting closed-source large language models (LLMs) with advanced reasoning strategies such as chain-of-thought~\citep{li2025alphasqlzeroshottexttosqlusing,zhai2025excotoptimizingreasoningtexttosql,pourreza2024chasesqlmultipathreasoningpreference}; and (iii) Reinforcement Learning (RL)-based methods that optimize model reasoning with algorithms such as PPO and GRPO~\citep{pourreza2025reasoningsqlreinforcementlearningsql,ma2025sqlr1trainingnaturallanguage,yao2025arctictext2sqlr1simplerewardsstrong,dai2025reexsqlreasoningexecutionawarereinforcement}. In particular, RL methods leverage final SQL execution results as reward signals for policy optimization. However, current practices treat execution feedback merely as scalar rewards, wasting rich tool information and leaving static LLMs unable to adapt their reasoning dynamically.

Recently, Multi-turn Tool-Integrated Reasoning (MTIR) has emerged as a promising paradigm for enhancing the reasoning capabilities of LLMs. By interleaving model reasoning with external tools---such as search engines, Python interpreters, and SQL executors---MTIR enables LLMs to overcome intrinsic limitations in computation, retrieval, and structured manipulation. Several studies have demonstrated the benefits of RL-based optimization in tool use: Search-R1~\citep{jin2025searchr1trainingllmsreason} explores dynamic tool invocation for question answering, while ToRL~\citep{li2025torlscalingtoolintegratedrl} and Effective CIR~\citep{bai2025effectivecodeintegratedreasoning} design tailored RL recipes for mathematical reasoning. More recently, the ReEx-SQL~\citep{dai2025reexsqlreasoningexecutionawarereinforcement} framework extends TIR to Text-to-SQL, providing early evidence of its potential in structured query tasks.

Despite this progress, three fundamental challenges remain. On the \textbf{tool side}, SQL-oriented MTIR remains underexplored; existing efforts often rely on ad-hoc token mechanisms for tool invocation, limiting interoperability, extensibility, and compatibility with diverse database operations~\citep{jin2025searchr1trainingllmsreason,dai2025reexsqlreasoningexecutionawarereinforcement}. On the \textbf{RL side}, dominant methods such as GRPO suffer from reward collapse and difficulty in modeling long-horizon dependencies, leading to instability in multi-turn tool interaction. On the \textbf{framework side}, current Text-to-SQL studies adopt heterogeneous, non-standardized implementations that lack modularity and generality~\citep{dai2025reexsqlreasoningexecutionawarereinforcement,ma2025sqlr1trainingnaturallanguage,yao2025arctictext2sqlr1simplerewardsstrong,gajjar2025cognisqlr1zerolightweightreinforcedreasoning}.

\begin{figure}[t]
    \centering
    \includegraphics[width=1\textwidth]{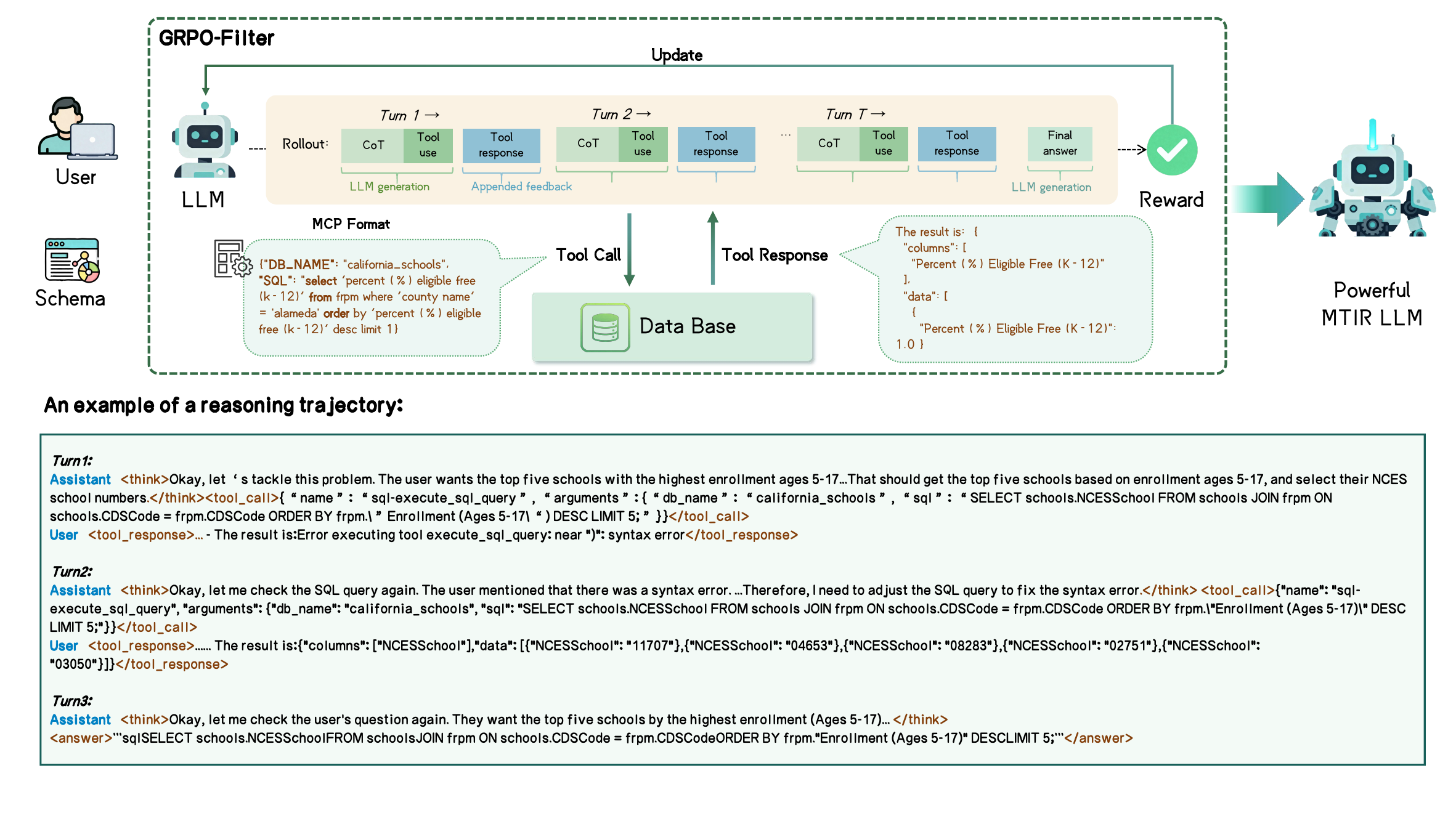}
    \caption{Overview of the MTIR-SQL framework. The framework integrates multi-turn reasoning with execution feedback and extends GRPO with trajectory filtering to enable dynamic correction and stable training, thereby enhancing SQL generation accuracy in complex scenarios.}
    \label{fig_1}
\end{figure}

To address these issues, we propose \textbf{MTIR-SQL}, a reinforcement learning framework for Multi-turn Tool-Integrated Reasoning in Text-to-SQL (Figure~\ref{fig_1}). MTIR-SQL extends GRPO to handle complex multi-turn interactions and introduces two key modifications: (i) a trajectory filtering mechanism to discard invalid rollouts and (ii) the removal of KL regularization to mitigate distributional collapse during training. Built on top of RL-Factory~\citep{chai2025rlfactoryplugandplayreinforcementlearning} with standardized MCP-compatible tool invocation, our framework ensures extensibility and interoperability.

Our contributions are summarized as follows:
\begin{itemize}
    \item \textbf{MTIR-SQL Framework.} We introduce a novel RL framework for Text-to-SQL that enables LLMs to reason interactively and directly optimize via SQL execution feedback. It incorporates retrieval-based token masking for stable training and supports multi-turn iterative reasoning and execution verification.
    \item \textbf{GRPO Extensions.} We extend GRPO with SQL execution rollout expansion and trajectory filtering to stabilize training in multi-turn tool-use scenarios, effectively mitigating reward collapse.
    \item \textbf{Strong Empirical Results.} On the BIRD dataset, MTIR-SQL trained on Qwen-3-4B achieves a 16\% absolute improvement over baselines, matching the performance of recent 7B-coder models. It attains competitive execution accuracy, demonstrating its robustness and effectiveness.
\end{itemize}

\section{Related Work}
\subsection{RL for Tool-Integrated Reasoning}

Tool-Integrated Reasoning (TIR) has emerged as a key paradigm for augmenting large language models by enabling interaction with external tools and APIs~\citep{zhang2025landscapeagenticreinforcementlearning}. Early work focused on single-turn tool invocation through supervised fine-tuning approaches, demonstrating effectiveness across domains, including mathematical reasoning, code generation~\citep{mai2025agentrlscalinglaw}, and search integration~\citep{jin2025searchr1trainingllmsreason}. Multi-turn TIR enables iterative reasoning through sequential tool interactions, where models repeatedly generate tool calls, execute it, and refine based on results~\citep{mai2025agentrlscalinglaw,shang2025rstar2agentagenticreasoningtechnical,fu2025srftsinglestagemethodsupervised,yue2025promotingefficientreasoningverifiable,lin2025restreshapingtokenlevelpolicy,wang2025agentflyextensiblescalablereinforcement,zeng2025reinforcingmultiturnreasoningllm,zhao2025muarlmultiturnuserinteractingagent}. However, training stability remains a critical challenge due to distribution drift from external tool outputs and error accumulation across reasoning rounds, often leading to training instability and entropy collapse. Recent work like SimpleTIR~\citep{xue2025simpletirendtoendreinforcementlearning} addresses these issues by filtering empty rounds in multi-turn reasoning, achieving state-of-the-art performance on mathematical tasks. Despite these advances, optimizing Multi-turn TIR for complex tasks remains challenging~\citep{lin2025understandingtoolintegratedreasoning,dong2025toolstarempoweringllmbrainedmultitool,dong2025agenticreinforcedpolicyoptimization,yu2025dapoopensourcellmreinforcement}. We apply recent MTIR advancements, including filtering and handling invalid turns, to the Text-to-SQL domain, improving execution feedback management, multi-table relationship handling, and ensuring SQL semantic correctness across iterations.
\subsection{Text-to-SQL}
Text-to-SQL aims to automatically convert natural language questions into executable SQL query statements, enabling natural language interfaces for databases. The field has evolved through three main paradigms: supervised fine-tuning methods that train specialized models on domain-specific datasets~\citep{li2024codesbuildingopensourcelanguage,li2025omnisqlsynthesizinghighqualitytexttosql,yang2024synthesizingtexttosqldataweak,qin2025routerobustmultitasktuning}, using closed-source large models with prompt engineering and chain-of-thought reasoning to handle complex multi-table joins and nested queries~\citep{li2025alphasqlzeroshottexttosqlusing,zhai2025excotoptimizingreasoningtexttosql,pourreza2024chasesqlmultipathreasoningpreference,pourreza2025reasoningsqlreinforcementlearningsql,lyu2025sqlo1selfrewardheuristicdynamic,pourreza2023dinsqldecomposedincontextlearning,xie2024magsqlmultiagentgenerativeapproach,cao2024rslsqlrobustschemalinking}, and reinforcement learning approaches using algorithms such as GRPO to enhance reasoning capabilities and generalization~\citep{dai2025reexsqlreasoningexecutionawarereinforcement}. Despite recent advancements, current RL-based methods exhibit significant limitations. They rely on static context during generation and lack mechanisms for validating or correcting intermediate reasoning steps, resulting in errors that cannot be self-corrected~\citep{ma2025sqlr1trainingnaturallanguage,yao2025arctictext2sqlr1simplerewardsstrong,gajjar2025cognisqlr1zerolightweightreinforcedreasoning}. Execution feedback is treated as a reward signal rather than dynamically integrated, hindering the model's ability to adapt to complex scenarios. The challenge persists in incorporating execution feedback while managing database results, multi-table relationships, and ensuring SQL semantic correctness. To address these issues, we introduce Multi-turn Tool-Integrated Reasoning in the Text-to-SQL domain, enabling the model to improve performance through iterative use of external tools.
\begin{figure}[h]
    \centering
    \includegraphics[width=1\textwidth]{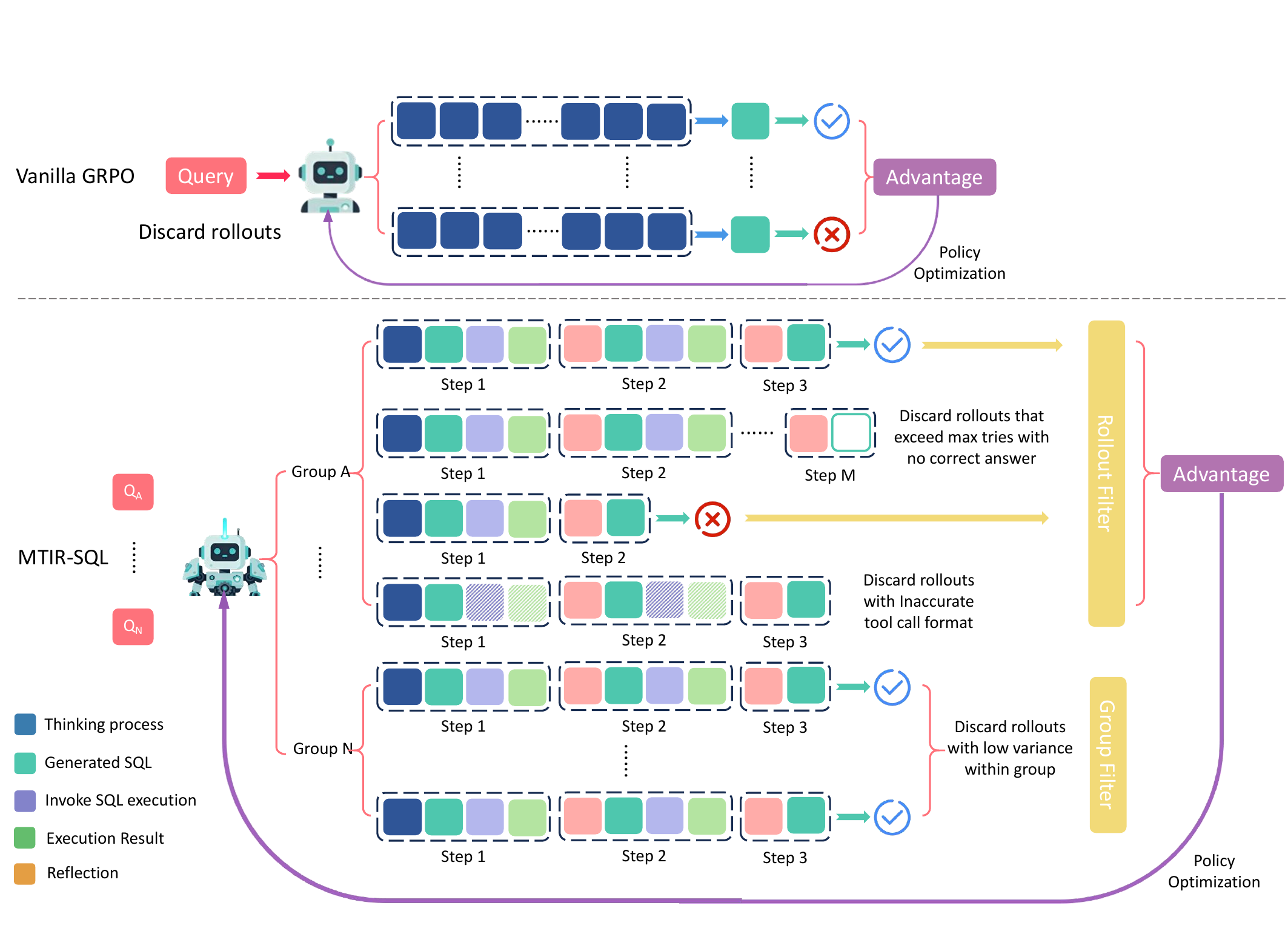}
    \caption{Compared to vanilla GRPO, our framework removes the KL constraint, introduces quality-aware rollout filtering, and extends to multi-turn reasoning with SQL execution feedback for more stable and accurate policy optimization.}
    \label{fig:example}    
\end{figure}
\section{Methodology}
We propose an SQL-integrated reinforcement learning framework with GRPO-Filter, which combines unconstrained optimization, selective rollout filtering, and multi-turn reasoning to improve decision-making. The model dynamically interacts with SQL execution, refining its output through iterative feedback. A reward mechanism focused on format, execution, and result correctness guides the generation of high-quality SQL queries.
\subsection{SQL-Integrated RL with GRPO-Filter}
We formulate the reinforcement learning framework with SQL execution tool $\mathcal{E}$ as follows:
\begin{equation}
\max_{\pi_\theta} \mathbb{E}_{x \sim D, y \sim \pi_\theta(\cdot|x;\mathcal{E})} [r_\phi(x,y)],
\end{equation}
where $\pi_\theta$ is the LLM policy and $r_\phi$ is the reward function. Unlike prior reinforcement learning methods that primarily rely on the policy LLM $\pi_\theta(\cdot | x)$ to generate rollout sequences, our framework explicitly incorporates SQL execution-guided reasoning via $\pi_\theta(\cdot | x;\mathcal{E})$, which can be formulated as $\pi_\theta(\cdot | x) \otimes \mathcal{E}$, where $\otimes$ denotes the interleaved SQL generation and execution feedback. This enables more effective decision-making in SQL generation tasks by leveraging real-time execution results to guide the model's reasoning process.

Our approach introduces \textbf{GRPO-Filter}, an enhanced variant of Group Relative Policy Optimization (GRPO) specifically designed for complex multi-turn interactive scenarios. GRPO-Filter incorporates three key innovations:

\textbf{Unconstrained Optimization:} Unlike standard GRPO, we remove the KL divergence constraint between the policy and reference model, allowing for more flexible policy updates:
\begin{equation}
\mathcal{L}_{\textnormal{GRPO-Filter}} = -\mathbb{E}_{(x,y) \sim \mathcal{D}} \left[ \frac{\pi_\theta(y|x;R)}{\pi_{\textnormal{ref}}(y|x;R)} \cdot A(x,y) \right],
\end{equation}
where $A(x,y)$ represents the advantage function, eliminating the traditional KL penalty term $\beta \cdot \text{KL}(\pi_\theta || \pi_{ref})$.

\textbf{Selective Rollout Filtering:} We implement a quality-aware filtering mechanism during rollout generation, where only high-quality trajectories satisfying predefined criteria are retained for training:
\begin{equation}
\mathcal{T}_{\textnormal{filtered}} = \{(x,y) \in \mathcal{T}_\textnormal{rollout} : \mathcal{F}(x,y) > \tau\},
\end{equation}
where $\mathcal{F}(\cdot)$ is the filtering function and $\tau$ is the quality threshold.

\textbf{Multi-turn Extension:} GRPO-Filter extends the original framework to handle complex multi-turn interactions by maintaining conversation context and enabling iterative reasoning across multiple dialogue turns:
\begin{equation}
\pi_\theta(y_t | x, h_{<t}; R) = \pi_\theta(y_t | \text{concat}(x, h_{<t}); R),
\end{equation}
where $h_{<t}$ represents the conversation history up to turn $t$, and $y_t$ is the response at turn $t$.

This multifaceted approach allows GRPO-Filter to effectively optimize policies for reasoning-intensive tasks while maintaining training stability and improving sample efficiency through selective learning from high-quality experiences.
\subsection{Interaction with SQL Execution Environment}
The integration of SQL and its execution interface with large language models (LLMs), which are capable of comprehending and generating query intentions, can significantly enhance the automation of complex database operations. In an LLM-based SQL tool invocation environment, the system should exhibit human-like interactive and reasoning behaviors. These behaviors include generating syntactically correct and logically sound SQL queries from natural language questions, invoking database execution interfaces at appropriate moments, and executing queries safely. Additionally, the system should carefully interpret query results, verify their correctness, and refine subsequent problem decomposition or query generation strategies based on feedback. This capability is cultivated through guiding the model via multi-turn interaction and reflective learning with the SQL execution environment. Detailed prompt of sql Execution can be found in Appendix \ref{system}.  

With the support of SQL tools, the model dynamically incorporates database query results into the reasoning process through multi-turn execution, as illustrated in Figure \ref{fig_1}. Unlike conventional methods that generate a complete SQL query until an end-of-sequence (EOS) token is produced, our approach constructs the full reasoning trajectory through continuous interaction with the SQL execution environment. The first interaction begins with a system prompt followed by the user's question, with detailed content available in Appendix \ref{system}. The model, acting as the assistant, generates an initial response until it outputs an EOS token. If no SQL tool call is detected, the process terminates. When an SQL query is identified, the environment service extracts and safely executes it, then appends the execution result to the dialogue context in the user role. The model subsequently continues its reasoning as the assistant based on the updated context, producing the next turn of response. This multi-turn process iterates until the model returns a final answer or a maximum number of turns, denoted as $T$, is reached. Detailed content can be found in Appendix \ref{app:algr1}.

\subsection{Reward Design}
To optimize policy effectively, we introduce a streamlined reward mechanism that focuses on critical elements of SQL query quality. This framework incorporates three key factors—syntax validity, execution feasibility, and semantic precision—each providing distinct guidance to ensure the model generates SQL queries that are syntactically correct, executable, and semantically meaningful.
\paragraph{Format Reward.} 
We guide the model to maintain a specific sequence of tags, ensuring a structured response. The response should follow a strict order: starting with \texttt{<think>...</think>}, followed optionally by \texttt{<tool\_call>...</tool\_response>}, and concluding with the \texttt{<answer>...</answer>} tag. Additionally, all tools must be used within \texttt{<tool\_call>...</tool\_call>} and \texttt{<tool\_response>...</tool\_response>} tags to maintain a standardized flow.

\begin{equation}
R_f =
\begin{cases}
0.1, & \text{if the format is correct}, \\
-0.1, & \text{if the format is incorrect}.
\end{cases}
\end{equation}

\paragraph{Execution Reward.}
This reward evaluates the syntactic correctness and executability of the generated SQL. It prevents the model from producing invalid or overly complex queries. If the SQL statement fails to execute, the model will not receive subsequent rewards. Furthermore, the execution time is constrained to discourage the generation of unnecessarily complex queries:
\begin{equation}
R_e =
\begin{cases}
0.1, & \text{if the SQL query is executable}, \\
0, & \text{if the format is incorrect}, \\
-0.1, & \text{if the SQL query is not executable}.
\end{cases}
\end{equation}

\paragraph{Result Reward.}
The correctness of query results is a crucial measure of semantic fidelity. To encourage faithful reasoning, we design the result reward to strongly differentiate between correct and incorrect outputs:
\begin{equation}
R_r =
\begin{cases}
1, & \text{if the query result is correct}, \\
0, & \text{if the format is incorrect or the SQL is not executable}, \\
-1, & \text{if the query result is incorrect}.
\end{cases}
\end{equation}

\begin{table}[t]
\centering
\caption{Comparison between MTIR-SQL and other models (Under 10B parameters)}
\label{tab:comparison_under_10b}
\begin{tabular}{lcc}
\toprule
\textbf{Model} & \textbf{Size} & \textbf{BIRD Dev (EX\%)} \\
\midrule
Granite-8B-Code-Instruct \cite{mishra2024granitecodemodelsfamily} & 8B & 27.6 \\
Granite-31.8B-Instruct\cite{mishra2024granitecodemodelsfamily} & 8B & 36.0 \\
OpenCoder-8B-Instruct \cite{huang2025opencoderopencookbooktoptier} & 8B & 37.5 \\
Meta-Llama-3.1B-Instruct \cite{grattafiori2024llama3herdmodels} & 8B & 42.0 \\
DPSK-Coder-6.7B-Instruct \cite{guo2024deepseekcoderlargelanguagemodel} & 6.7B & 43.1 \\
Qwen2.5-Coder-3B-Instruct \cite{hui2024qwen25codertechnicalreport} & 3B & 48.1 \\
Qwen2.5-Coder-7B-Instruct \cite{hui2024qwen25codertechnicalreport} & 7B & 50.9 \\
SFT Code5-7B \cite{li2024codesbuildingopensourcelanguage} & 7B & 57.17 \\
Think2SQL-7B \cite{papicchio2025think2sqlreinforcellmreasoning} & 7B & 56.1 \\
SQL-R1 + Qwen2.5-Coder-3B \cite{ma2025sqlr1trainingnaturallanguage} & 3B & 54.6 \\
SQL-R1 + Qwen2.5-Coder-7B \cite{ma2025sqlr1trainingnaturallanguage} & 7B & 63.1 \\
CogniSQL-R1-Zero + Qwen2.5-Coder-7B\cite{gajjar2025cognisqlr1zerolightweightreinforcedreasoning} & 7B & 59.17 \\
\midrule
\textbf{MTIR-SQL + Qwen3-4B(Ours)} & \textbf{4B} & \textbf{64.4} \\
\bottomrule
\end{tabular}
\end{table}

\section{Experiments}
\subsection{Experimental Setup}
\textbf{Datasets.}
We train and evaluate our model on two Text-to-SQL benchmarks, SPIDER~\citep{yu2019SPIDERlargescalehumanlabeleddataset} and BIRD~\citep{li2023llmservedatabaseinterface}, which assess different aspects of the task. SPIDER is a large-scale, cross-domain benchmark focused on SQL complexity, with 10,181 questions and 5,693 unique queries across 200 databases. BIRD addresses real-world scenarios, featuring 12,751 question-SQL pairs on 95 large-scale databases with "dirty" data and evaluating both accuracy and efficiency.To ensure both training efficiency and SQL generation accuracy, this study follows the principles of "high quality, executable, and low redundancy" for data filtering and optimization. For the training of the BIRD and SPIDER benchmarks, we prioritize execution validity checks. Batch execution of reference SQL queries revealed that some samples returned empty results, which, if used for RL training, would fail to provide valid reward signals and could lead to learning biases or "reward hacking."

\textbf{Baselines.}
We compare our MTIR-SQL framework against two primary categories of baseline methods. For supervised fine-tuning, we evaluate Qwen2.5-Coder-7B-Instruct~\citep{hui2024qwen25codertechnicalreport} , a state-of-the-art code generation model fine-tuned on Text-to-SQL datasets using standard cross-entropy loss. For reinforcement learning without tool integration, we implement GRPO on the Qwen3-4B model, using execution accuracy as the reward signal to optimize SQL generation through policy gradient methods. Both baselines use identical training procedures and computational budgets as our proposed framework but lack access to intermediate execution feedback during generation, allowing us to isolate the contribution of Multi-turn Tool-Integrated Reasoning.

\textbf{Experimental Details.}
We conduct experiments using the Qwen3-Instruct model. During training and inference, we adopt database prompts from CodeS~\citep{li2024codesbuildingopensourcelanguage} and SQL-R1~\citep{ma2025sqlr1trainingnaturallanguage}, which provide curated schema components, values, and metadata, and have demonstrated competitive performance on the BIRD benchmark. We employ algorithms such as PPO and GRPO within the RL-Factory framework~\citep{chai2025rlfactoryplugandplayreinforcementlearning}.The training configuration uses a batch size of 64 and a learning rate of 1e-6. During the rollout phase, we sample 5 outputs for each input at temperature T = 0.6, set the maximum sequence length to 8192, and the maximum number of interactions to N = 6. During inference, we apply greedy decoding (T = 0.0). We use SQLite as the SQL executor to obtain execution feedback. The feedback includes column headers and cell values for up to 10 rows. All experiments are conducted on a system equipped with 8 NVIDIA A100 GPUs.

\begin{table}[t]
\centering
\caption{Comparison between MTIR-SQL and other models (10B--30B parameters)}
\label{tab:comparison_10b_30b}
\begin{tabular}{lcc}
\toprule
\textbf{Model} & \textbf{Size} & \textbf{BIRD Dev (EX\%)} \\
\midrule
Granite-20B-Code-Instruct \citep{mishra2024granitecodemodelsfamily} & 20B & 34.0 \\
Starcoder-15B-Instruct \citep{lozhkov2024starcoder2stackv2} & 15B & 38.5 \\
DPSK-Coder-V2-Instruct (MoE) \citep{deepseekai2024deepseekcoderv2breakingbarrierclosedsource} & 16B & 44.6 \\
Codestral-22B \citep{mistral_codestral} & 22B & 52.7 \\
Qwen2.5-14B-Instruct \citep{hui2024qwen25codertechnicalreport} & 14B & 56.7 \\
SFT Code5-15B \citep{li2024codesbuildingopensourcelanguage} & 15B & 58.47 \\
\midrule
\textbf{MTIR-SQL + Qwen3-4B (Ours)} & \textbf{4B} & \textbf{64.4} \\
\bottomrule
\end{tabular}
\end{table}

\begin{table}[t]
\centering
\caption{Comparison between MTIR-SQL and large-scale models}
\label{tab:comparison_large}
\begin{tabular}{lcc}
\toprule
\textbf{Model} & \textbf{Size} & \textbf{BIRD Dev (EX\%)}\\
\midrule
\multicolumn{3}{l}{\textit{Open-source Models}} \\
Granite-34B-Code-Instruct \citep{mishra2024granitecodemodelsfamily} & 34B & 33.8 \\
Codestral Base \citep{li2023llmservedatabaseinterface} & 175B & 34.35 \\
Mixtral-8x7B-Instruct (MoE) \citep{jiang2024mixtralexperts} & 47B & 35.3 \\
\midrule
\multicolumn{3}{l}{\textit{Proprietary Models}} \\
ChatGPT Baseline \citep{li2023llmservedatabaseinterface} & -- & 37.22 \\
Claude-3 \citep{li2023llmservedatabaseinterface} & -- & 42.70 \\
GPT-4 Baseline \citep{li2023llmservedatabaseinterface} & -- & 46.35 \\
ChatGPT-4 Baseline \citep{deepseekai2024deepseekcoderv2breakingbarrierclosedsource} & -- & 49.2 \\
Mistral Baseline \citep{li2023llmservedatabaseinterface} & 123B & 53.52 \\
DeepSeek Baseline \citep{li2023llmservedatabaseinterface} & 236B & 56.13 \\
SuperSQL (NLSQL-1360) \citep{Li_2024} & -- & 58.50 \\
ChatGPT + CoT \citep{li2023llmservedatabaseinterface} & -- & \textbf{64.64} \\
\midrule
\textbf{MTIR-SQL + Qwen3-4B (Ours)} & 4B & 64.4 \\
\bottomrule
\end{tabular}
\end{table}

\begin{table}[t]
\centering
\caption{Performance comparison of reasoning paradigms on benchmarks with pass@1.}
\label{tab:reasoning_paradigms}
\begin{tabular}{@{}lcccc@{}}
\toprule
\textbf{Reasoning Paradigm} & \textbf{Training Type} & \textbf{BIRD Dev} & \textbf{SPIDER Dev} & \textbf{SPIDER Test} \\
 & & \textbf{EX (\%)} & \textbf{EX (\%)} & \textbf{EX (\%)} \\
\midrule
Direct Output & -- & 46.9 & 69.2 & 70.8\\
Standard Reasoning & -- & 48.1 & 72.5 & 72.9 \\
Tool-Integrated Reasoning & -- & 47.6 & 71.1 & 73.6 \\
\midrule
Standard Reasoning & GRPO & 58.9 & 78.2 & 79.1 \\
Multi-turn TIR & PPO & 58.2& 77.2 & 79.2 \\
Multi-turn TIR & GRPO & 60.3& 80.1 & 81.4 \\
\textbf{Multi-turn TIR} & \textbf{GRPO-Filter} & \textbf{63.1} & \textbf{82.4} & \textbf{83.4} \\
\bottomrule
\end{tabular}
\end{table}

\subsection{Main Result}
\textbf{Performance on Main Benchmarks.}
In the first set of comparisons (Tables~\ref{tab:comparison_under_10b}, \ref{tab:comparison_10b_30b}, and \ref{tab:comparison_large}), we evaluate MTIR-SQL against models with different parameter sizes, ranging from under 10B to large-scale models. As shown in Table~\ref{tab:comparison_under_10b}, our MTIR-SQL model (4B) achieves 64.4\% on the BIRD Dev set, which is significantly higher than the other models under 10B parameters. The next best-performing model, SFT Code5-7B, achieves 57.17\%, marking a clear advantage for our model despite its smaller size.

In Table~\ref{tab:comparison_10b_30b}, MTIR-SQL (4B) continues to outperform all models in the 10B-30B parameter range, where the best result from SFT Code5-15B is 58.47\%. This demonstrates the efficiency of MTIR-SQL in achieving competitive performance even with a smaller model size. Similarly, in Table~\ref{tab:comparison_large}, we compare MTIR-SQL to large-scale models, including both open-source and proprietary models. Despite the large model sizes (up to 236B parameters), MTIR-SQL (4B) achieves comparable performance to models such as ChatGPT + CoT (64.64\%) and SuperSQL (58.50\%), showing the robustness and scalability of our approach in handling complex Text-to-SQL tasks.

\textbf{Performance on Reasoning Paradigms.}
In Table~\ref{tab:reasoning_paradigms}, we evaluate the performance of different reasoning paradigms on Text-to-SQL benchmarks, specifically focusing on Pass@1 performance across the SPIDER and BIRD datasets. The results highlight the effectiveness of multiturn tool-integrated reasoning.
Among the reasoning paradigms, multi-turn TIR with GRPO-Filter leads to the highest performance on both the BIRD and SPIDER benchmarks. The BIRD Dev score of 63.1\% represents a significant improvement over standard reasoning and tool-integrated reasoning paradigms, which score 48.1\% and 47.6\%, respectively. In SPIDER Dev and Test, multi-turn TIR with GRPO filter also excels, achieving 82.4\% and 83.4\%, respectively, marking a clear advantage over other paradigms.

This reinforces the importance of incorporating execution feedback through multi-turn reasoning for enhancing performance in real-world Text-to-SQL tasks, particularly when dealing with complex databases like SPIDER and BIRD.

\subsection{Ablation Study}
\begin{figure}[h]
    \centering
    \includegraphics[width=1\textwidth]{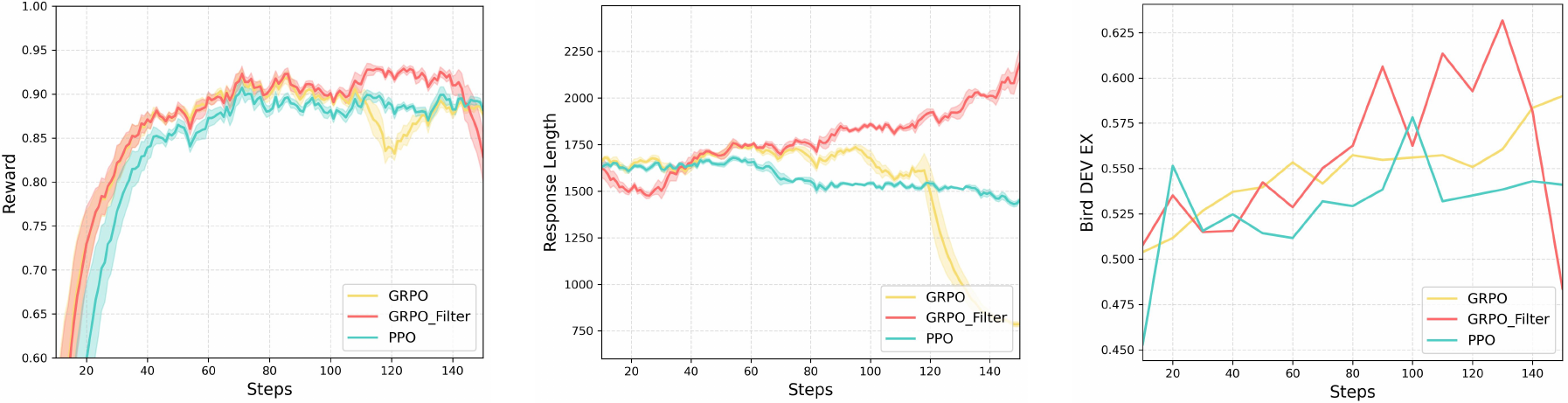}
    \caption{Comparing the impact of different RL Methods on training and performance.}
    \label{fig:RL Methods}    
\end{figure}

\textbf{Ablation Study of RL Methods.}
To assess the effectiveness of MTIR-SQL, we conducted comparisons against PPO, GRPO, and our improved GRPO-Filter using the Qwen3-4B model. As illustrated in Figure~\ref{fig:RL Methods} and summarized in Table~\ref{tab:reasoning_paradigms}, GRPO converges more rapidly than PPO due to the absence of a critic warm-up phase, but it often suffers from reward collapse in later training stages. PPO, in contrast, provides greater stability but at the cost of slower convergence. Crucially, GRPO-Filter addresses these limitations by selectively filtering low-quality rollouts and removing the KL constraint, thereby stabilizing multi-turn training while achieving substantial performance gains. This demonstrates that our modifications are not merely incremental but essential for enabling robust reinforcement learning in execution-aware Text-to-SQL tasks.  

\begin{figure}[h]
    \centering
    \includegraphics[width=1\textwidth]{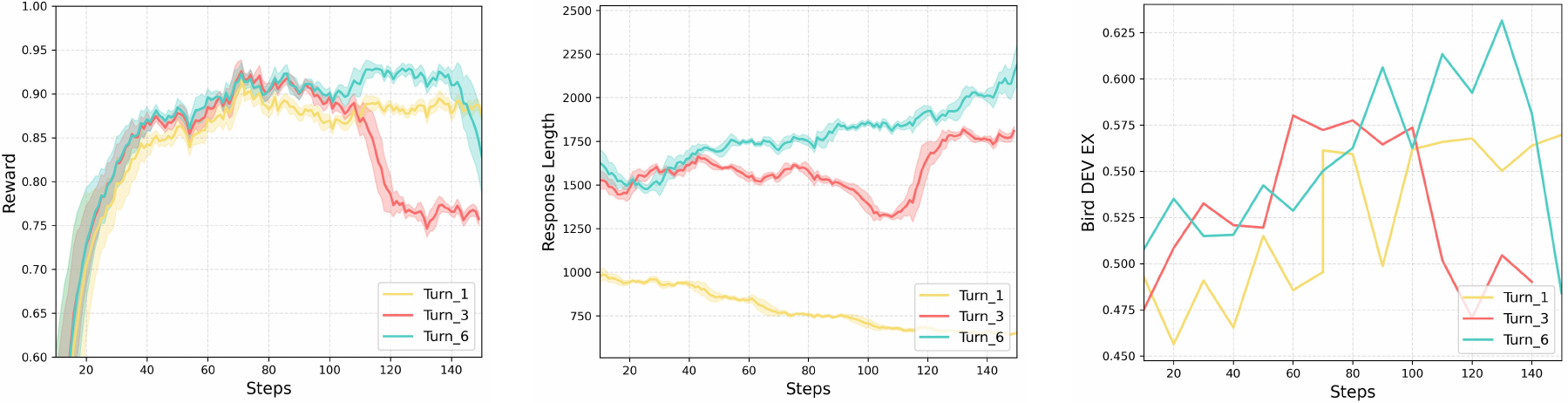}
    \caption{Comparing the impact of different max turns on training and performance.}
    \label{fig:Max Turn}    
\end{figure}
\textbf{Ablation Study on Max Turns.}
We further analyzed the impact of maximum tool calling turns by conducting experiments with settings of 1, 3, and 6. The training curves are shown in Figure~\ref{fig:Max Turn}, and the quantitative results are summarized in Table~\ref{tab:reasoning_paradigms}. The main findings are as follows: (1) Larger Max Turns generally lead to higher final rewards and stronger overall performance. More turns provide the model with additional opportunities to explore, optimize responses, and execute complex tasks; (2) However, excessive turns (such as 6) may also result in training instability, occasionally causing reward saturation or collapse phenomena; (3) Although Max Turns = 1 demonstrates faster convergence, due to severely limited interaction flexibility, there exists a gap between the final performance and optimal values.

\begin{figure}[h]
    \centering
    \begin{minipage}{0.45\textwidth}
        \centering
        \captionof{table}{Ablation of Reward Components for MTIR-SQL on BIRD Dev.}
        \begin{tabular}{ccc}
        \toprule
        \textbf{Reward Components} & \textbf{BIRD Dev (EX \%)} \\
        \midrule
        MTIR-SQL & 63.1 \\
        w/o $R_{\text{format}}$ & 62.3 $\downarrow$ (0.8) \\
        w/o $R_{\text{exec}}$ & 59.4 $\downarrow$ (3.9) \\
        w/o $R_{\text{result}}$ & 58.8 $\downarrow$ (4.3) \\
        \bottomrule
        \end{tabular}
    \end{minipage}
    \hfill
    \begin{minipage}{0.45\textwidth}
        \centering
        \includegraphics[width=\textwidth]{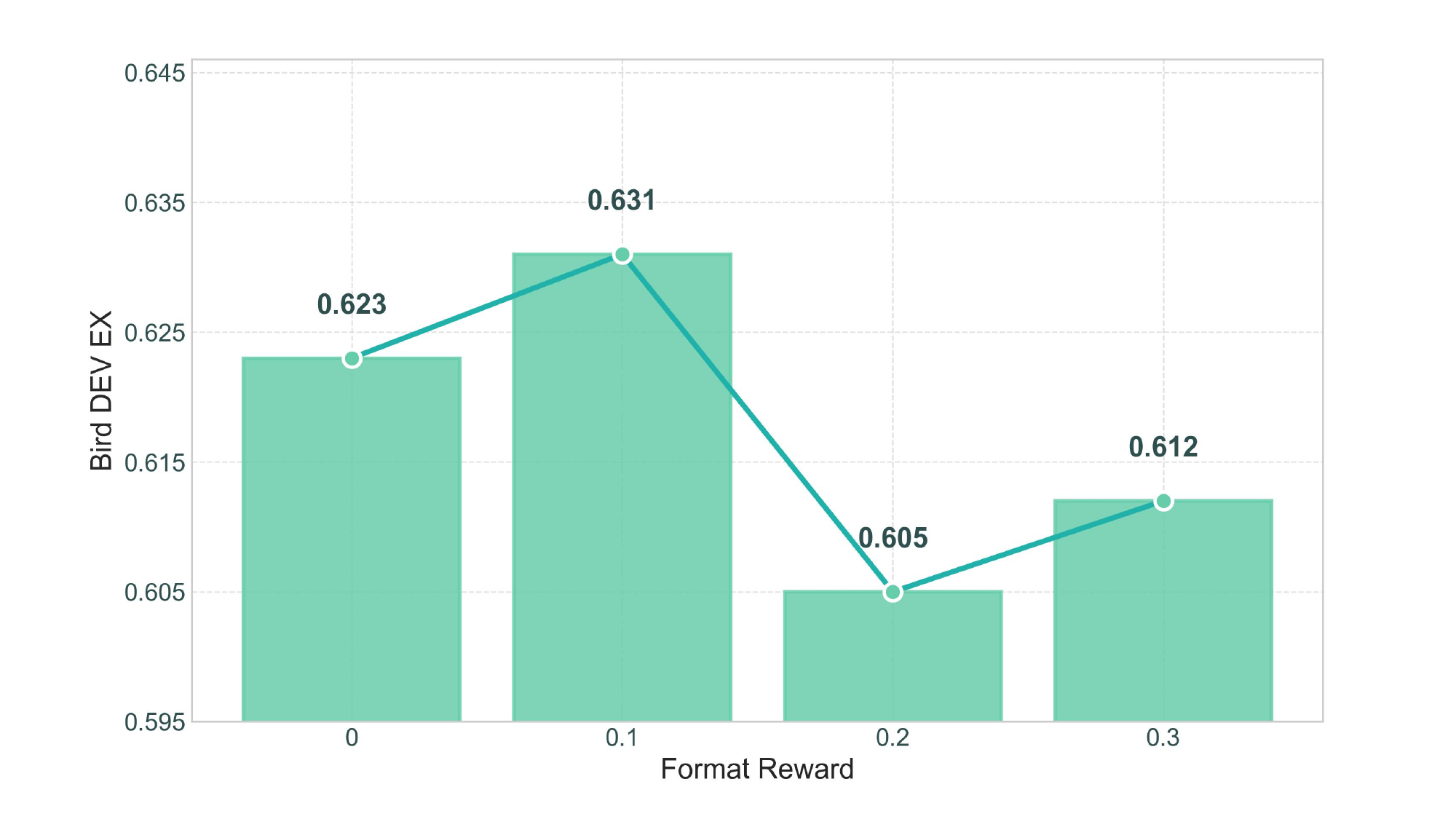}
        \captionof{figure}{Ablation of Reward Components for MTIR-SQL on BIRD Dev Reward.}
        \label{reward}
    \end{minipage}
\end{figure}

\textbf{Ablation Study on Reward Design.}
This ablation study evaluates the impact of removing each reward component on the model’s performance using the BIRD Development set:

\begin{itemize}
    \item \textbf{$R_{\text{f}}$ (Exploration Reward):}  
    Removing the exploration reward shows that a moderate increase in its value (from 0 to 0.1) improves performance from 62.3\% to 63.1\%, a 0.8\% increase. However, further increases (to 0.2) cause a significant drop in performance, from 63.1\% to 60.5\% (a 4.1\% decline), indicating that excessive exploration rewards can negatively impact performance. A slight recovery is observed at 0.3 (60.5\% to 61.2\%, a 1.2\% increase), suggesting that a balanced exploration reward is beneficial, but it should not be overemphasized.
    
    \item \textbf{$R_{\text{e}}$ (Execution Reward):}  
    Removing the execution reward results in the largest performance drop, from 63.1\% to 59.4\% (a 3.9\% decrease), highlighting its crucial role in the natural language-to-SQL conversion process. Without execution-based feedback, the model struggles to make accurate predictions.
    
    \item \textbf{$R_{\text{r}}$ (Result Reward):}  
    Excluding the result reward leads to a smaller decline in performance, from 63.1\% to 58.8\% (a 4.3\% drop), underlining its importance in ensuring the functional correctness of the model’s SQL queries.
\end{itemize}
In conclusion, removing any reward—particularly $R_{\text{e}}$—significantly hampers the model’s performance. This underscores the necessity of a balanced reward system that integrates execution feedback, exploration, and result accuracy for optimal performance.

\section{conclusion}
We propose \textbf{MTIR-SQL}, a novel reinforcement learning framework for complex multi-turn SQL generation tasks. MTIR-SQL's central innovation resides in its feedback-driven reasoning approach, where execution results inform subsequent reasoning iterations, creating a self-correcting mechanism that substantially improves generation stability and query accuracy. We extend GRPO with \textbf{trajectory filtering} to mitigate distribution drift and remove KL divergence constraints to enhance learning efficiency. Experimental results demonstrate MTIR-SQL's effectiveness: achieving \textbf{64.4\%} accuracy on BIRD-SQL and \textbf{84.6\%} execution accuracy on SPIDER with a 4B-parameter model, significantly outperforming baseline methods and advancing state-of-the-art in Text-to-SQL generation.

\bibliography{iclr2026_conference}

\begin{thebibliography}{49}
\providecommand{\natexlab}[1]{#1}
\providecommand{\url}[1]{\texttt{#1}}
\expandafter\ifx\csname urlstyle\endcsname\relax
  \providecommand{\doi}[1]{doi: #1}\else
  \providecommand{\doi}{doi: \begingroup \urlstyle{rm}\Url}\fi

\bibitem[Bai et~al.(2025)Bai, Min, Zhang, Chen, Zhao, Fang, Liu, Wang, and Wen]{bai2025effectivecodeintegratedreasoning}
Fei Bai, Yingqian Min, Beichen Zhang, Zhipeng Chen, Wayne~Xin Zhao, Lei Fang, Zheng Liu, Zhongyuan Wang, and Ji-Rong Wen.
\newblock Towards effective code-integrated reasoning, 2025.
\newblock URL \url{https://arxiv.org/abs/2505.24480}.

\bibitem[Cao et~al.(2024)Cao, Zheng, Fan, Zhang, Chen, and Bai]{cao2024rslsqlrobustschemalinking}
Zhenbiao Cao, Yuanlei Zheng, Zhihao Fan, Xiaojin Zhang, Wei Chen, and Xiang Bai.
\newblock {RSL-SQL: Robust Schema Linking in Text-to-SQL Generation}, 2024.
\newblock URL \url{https://arxiv.org/abs/2411.00073}.

\bibitem[Chai et~al.(2025)Chai, Yin, Xu, Yue, Jia, Xia, Wang, Jiang, Li, Dong, He, and Lin]{chai2025rlfactoryplugandplayreinforcementlearning}
Jiajun Chai, Guojun Yin, Zekun Xu, Chuhuai Yue, Yi~Jia, Siyu Xia, Xiaohan Wang, Jiwen Jiang, Xiaoguang Li, Chengqi Dong, Hang He, and Wei Lin.
\newblock Rlfactory: A plug-and-play reinforcement learning post-training framework for llm multi-turn tool-use, 2025.
\newblock URL \url{https://arxiv.org/abs/2509.06980}.

\bibitem[Dai et~al.(2025)Dai, Xie, Zhuang, Yang, Yang, Yang, Zhao, Chao, and Jiang]{dai2025reexsqlreasoningexecutionawarereinforcement}
Yaxun Dai, Wenxuan Xie, Xialie Zhuang, Tianyu Yang, Yiying Yang, Haiqin Yang, Yuhang Zhao, Pingfu Chao, and Wenhao Jiang.
\newblock Reex-sql: Reasoning with execution-aware reinforcement learning for text-to-sql, 2025.
\newblock URL \url{https://arxiv.org/abs/2505.12768}.

\bibitem[DeepSeek-AI et~al.(2024)DeepSeek-AI, Zhu, Guo, Shao, Yang, Wang, Xu, Wu, Li, Gao, Ma, Zeng, Bi, Gu, Xu, Dai, Dong, Zhang, Piao, Gou, Xie, Hao, Wang, Song, Chen, Xie, Guan, You, Liu, Du, Gao, Lu, Chen, Wang, Deng, Li, Zhao, Ruan, Luo, and Liang]{deepseekai2024deepseekcoderv2breakingbarrierclosedsource}
DeepSeek-AI, Qihao Zhu, Daya Guo, Zhihong Shao, Dejian Yang, Peiyi Wang, Runxin Xu, Y.~Wu, Yukun Li, Huazuo Gao, Shirong Ma, Wangding Zeng, Xiao Bi, Zihui Gu, Hanwei Xu, Damai Dai, Kai Dong, Liyue Zhang, Yishi Piao, Zhibin Gou, Zhenda Xie, Zhewen Hao, Bingxuan Wang, Junxiao Song, Deli Chen, Xin Xie, Kang Guan, Yuxiang You, Aixin Liu, Qiushi Du, Wenjun Gao, Xuan Lu, Qinyu Chen, Yaohui Wang, Chengqi Deng, Jiashi Li, Chenggang Zhao, Chong Ruan, Fuli Luo, and Wenfeng Liang.
\newblock Deepseek-coder-v2: Breaking the barrier of closed-source models in code intelligence, 2024.
\newblock URL \url{https://arxiv.org/abs/2406.11931}.

\bibitem[Dong et~al.(2025{\natexlab{a}})Dong, Chen, Li, Jin, Qian, Zhu, Mao, Zhou, Dou, and Wen]{dong2025toolstarempoweringllmbrainedmultitool}
Guanting Dong, Yifei Chen, Xiaoxi Li, Jiajie Jin, Hongjin Qian, Yutao Zhu, Hangyu Mao, Guorui Zhou, Zhicheng Dou, and Ji-Rong Wen.
\newblock Tool-star: Empowering llm-brained multi-tool reasoner via reinforcement learning, 2025{\natexlab{a}}.
\newblock URL \url{https://arxiv.org/abs/2505.16410}.

\bibitem[Dong et~al.(2025{\natexlab{b}})Dong, Mao, Ma, Bao, Chen, Wang, Chen, Du, Wang, Zhang, Zhou, Zhu, Wen, and Dou]{dong2025agenticreinforcedpolicyoptimization}
Guanting Dong, Hangyu Mao, Kai Ma, Licheng Bao, Yifei Chen, Zhongyuan Wang, Zhongxia Chen, Jiazhen Du, Huiyang Wang, Fuzheng Zhang, Guorui Zhou, Yutao Zhu, Ji-Rong Wen, and Zhicheng Dou.
\newblock Agentic reinforced policy optimization, 2025{\natexlab{b}}.
\newblock URL \url{https://arxiv.org/abs/2507.19849}.

\bibitem[Fu et~al.(2025)Fu, Chen, Chai, Wang, Tu, Yin, Lin, Zhang, Zhu, and Zhao]{fu2025srftsinglestagemethodsupervised}
Yuqian Fu, Tinghong Chen, Jiajun Chai, Xihuai Wang, Songjun Tu, Guojun Yin, Wei Lin, Qichao Zhang, Yuanheng Zhu, and Dongbin Zhao.
\newblock Srft: A single-stage method with supervised and reinforcement fine-tuning for reasoning, 2025.
\newblock URL \url{https://arxiv.org/abs/2506.19767}.

\bibitem[Gajjar et~al.(2025)Gajjar, Sikchi, Gautam, Hammons, and Jha]{gajjar2025cognisqlr1zerolightweightreinforcedreasoning}
Kushal Gajjar, Harshit Sikchi, Arpit~Singh Gautam, Marc Hammons, and Saurabh Jha.
\newblock Cognisql-r1-zero: Lightweight reinforced reasoning for efficient sql generation, 2025.
\newblock URL \url{https://arxiv.org/abs/2507.06013}.

\bibitem[Guo et~al.(2024)Guo, Zhu, Yang, Xie, Dong, Zhang, Chen, Bi, Wu, Li, Luo, Xiong, and Liang]{guo2024deepseekcoderlargelanguagemodel}
Daya Guo, Qihao Zhu, Dejian Yang, Zhenda Xie, Kai Dong, Wentao Zhang, Guanting Chen, Xiao Bi, Y.~Wu, Y.~K. Li, Fuli Luo, Yingfei Xiong, and Wenfeng Liang.
\newblock Deepseek-coder: When the large language model meets programming -- the rise of code intelligence, 2024.
\newblock URL \url{https://arxiv.org/abs/2401.14196}.

\bibitem[Hong et~al.(2025)Hong, Yuan, Zhang, Chen, Dong, Huang, and Huang]{hong2025nextgenerationdatabaseinterfacessurvey}
Zijin Hong, Zheng Yuan, Qinggang Zhang, Hao Chen, Junnan Dong, Feiran Huang, and Xiao Huang.
\newblock Next-generation database interfaces: A survey of llm-based text-to-sql, 2025.
\newblock URL \url{https://arxiv.org/abs/2406.08426}.

\bibitem[Huang et~al.(2025)Huang, Cheng, Liu, Hao, Song, Xu, Yang, Liu, Zhang, Chai, Yuan, Zhang, Fu, Liu, Zhang, Wang, Qi, Xu, and Chu]{huang2025opencoderopencookbooktoptier}
Siming Huang, Tianhao Cheng, J.~K. Liu, Jiaran Hao, Liuyihan Song, Yang Xu, J.~Yang, Jiaheng Liu, Chenchen Zhang, Linzheng Chai, Ruifeng Yuan, Zhaoxiang Zhang, Jie Fu, Qian Liu, Ge~Zhang, Zili Wang, Yuan Qi, Yinghui Xu, and Wei Chu.
\newblock Opencoder: The open cookbook for top-tier code large language models, 2025.
\newblock URL \url{https://arxiv.org/abs/2411.04905}.

\bibitem[Hui et~al.(2024)Hui, Yang, Cui, Yang, Liu, Zhang, Liu, Zhang, Yu, Lu, Dang, Fan, Zhang, Yang, Men, Huang, Zheng, Miao, Quan, Feng, Ren, Ren, Zhou, and Lin]{hui2024qwen25codertechnicalreport}
Binyuan Hui, Jian Yang, Zeyu Cui, Jiaxi Yang, Dayiheng Liu, Lei Zhang, Tianyu Liu, Jiajun Zhang, Bowen Yu, Keming Lu, Kai Dang, Yang Fan, Yichang Zhang, An~Yang, Rui Men, Fei Huang, Bo~Zheng, Yibo Miao, Shanghaoran Quan, Yunlong Feng, Xingzhang Ren, Xuancheng Ren, Jingren Zhou, and Junyang Lin.
\newblock Qwen2.5-coder technical report, 2024.
\newblock URL \url{https://arxiv.org/abs/2409.12186}.

\bibitem[Jiang et~al.(2024)Jiang, Sablayrolles, Roux, Mensch, Savary, Bamford, Chaplot, de~las Casas, Hanna, Bressand, Lengyel, Bour, Lample, Lavaud, Saulnier, Lachaux, Stock, Subramanian, Yang, Antoniak, Scao, Gervet, Lavril, Wang, Lacroix, and Sayed]{jiang2024mixtralexperts}
Albert~Q. Jiang, Alexandre Sablayrolles, Antoine Roux, Arthur Mensch, Blanche Savary, Chris Bamford, Devendra~Singh Chaplot, Diego de~las Casas, Emma~Bou Hanna, Florian Bressand, Gianna Lengyel, Guillaume Bour, Guillaume Lample, Lélio~Renard Lavaud, Lucile Saulnier, Marie-Anne Lachaux, Pierre Stock, Sandeep Subramanian, Sophia Yang, Szymon Antoniak, Teven~Le Scao, Théophile Gervet, Thibaut Lavril, Thomas Wang, Timothée Lacroix, and William~El Sayed.
\newblock Mixtral of experts, 2024.
\newblock URL \url{https://arxiv.org/abs/2401.04088}.

\bibitem[Jin et~al.(2025)Jin, Zeng, Yue, Yoon, Arik, Wang, Zamani, and Han]{jin2025searchr1trainingllmsreason}
Bowen Jin, Hansi Zeng, Zhenrui Yue, Jinsung Yoon, Sercan Arik, Dong Wang, Hamed Zamani, and Jiawei Han.
\newblock Search-r1: Training llms to reason and leverage search engines with reinforcement learning, 2025.
\newblock URL \url{https://arxiv.org/abs/2503.09516}.

\bibitem[Li et~al.(2024{\natexlab{a}})Li, Luo, Chai, Li, and Tang]{Li_2024}
Boyan Li, Yuyu Luo, Chengliang Chai, Guoliang Li, and Nan Tang.
\newblock The dawn of natural language to sql: Are we fully ready?
\newblock \emph{Proceedings of the VLDB Endowment}, 17\penalty0 (11):\penalty0 3318–3331, July 2024{\natexlab{a}}.
\newblock ISSN 2150-8097.
\newblock \doi{10.14778/3681954.3682003}.
\newblock URL \url{http://dx.doi.org/10.14778/3681954.3682003}.

\bibitem[Li et~al.(2025{\natexlab{a}})Li, Zhang, Fan, Xu, Chen, Tang, and Luo]{li2025alphasqlzeroshottexttosqlusing}
Boyan Li, Jiayi Zhang, Ju~Fan, Yanwei Xu, Chong Chen, Nan Tang, and Yuyu Luo.
\newblock Alpha-sql: Zero-shot text-to-sql using monte carlo tree search, 2025{\natexlab{a}}.
\newblock URL \url{https://arxiv.org/abs/2502.17248}.

\bibitem[Li et~al.(2024{\natexlab{b}})Li, Zhang, Liu, Fan, Zhang, Zhu, Wei, Pan, Li, and Chen]{li2024codesbuildingopensourcelanguage}
Haoyang Li, Jing Zhang, Hanbing Liu, Ju~Fan, Xiaokang Zhang, Jun Zhu, Renjie Wei, Hongyan Pan, Cuiping Li, and Hong Chen.
\newblock Codes: Towards building open-source language models for text-to-sql, 2024{\natexlab{b}}.
\newblock URL \url{https://arxiv.org/abs/2402.16347}.

\bibitem[Li et~al.(2025{\natexlab{b}})Li, Wu, Zhang, Huang, Zhang, Jiang, Wang, Zhang, Chen, Shi, Chen, and Li]{li2025omnisqlsynthesizinghighqualitytexttosql}
Haoyang Li, Shang Wu, Xiaokang Zhang, Xinmei Huang, Jing Zhang, Fuxin Jiang, Shuai Wang, Tieying Zhang, Jianjun Chen, Rui Shi, Hong Chen, and Cuiping Li.
\newblock Omnisql: Synthesizing high-quality text-to-sql data at scale, 2025{\natexlab{b}}.
\newblock URL \url{https://arxiv.org/abs/2503.02240}.

\bibitem[Li et~al.(2023)Li, Hui, Qu, Yang, Li, Li, Wang, Qin, Cao, Geng, Huo, Zhou, Ma, Li, Chang, Huang, Cheng, and Li]{li2023llmservedatabaseinterface}
Jinyang Li, Binyuan Hui, Ge~Qu, Jiaxi Yang, Binhua Li, Bowen Li, Bailin Wang, Bowen Qin, Rongyu Cao, Ruiying Geng, Nan Huo, Xuanhe Zhou, Chenhao Ma, Guoliang Li, Kevin C.~C. Chang, Fei Huang, Reynold Cheng, and Yongbin Li.
\newblock Can llm already serve as a database interface? a big bench for large-scale database grounded text-to-sqls, 2023.
\newblock URL \url{https://arxiv.org/abs/2305.03111}.

\bibitem[Li et~al.(2025{\natexlab{c}})Li, Zou, and Liu]{li2025torlscalingtoolintegratedrl}
Xuefeng Li, Haoyang Zou, and Pengfei Liu.
\newblock Torl: Scaling tool-integrated rl, 2025{\natexlab{c}}.
\newblock URL \url{https://arxiv.org/abs/2503.23383}.

\bibitem[Lin \& Xu(2025)Lin and Xu]{lin2025understandingtoolintegratedreasoning}
Heng Lin and Zhongwen Xu.
\newblock Understanding tool-integrated reasoning, 2025.
\newblock URL \url{https://arxiv.org/abs/2508.19201}.

\bibitem[Lin et~al.(2025)Lin, Wang, Cao, Chai, Yin, Lin, and He]{lin2025restreshapingtokenlevelpolicy}
Zihan Lin, Xiaohan Wang, Jie Cao, Jiajun Chai, Guojun Yin, Wei Lin, and Ran He.
\newblock Rest: Reshaping token-level policy gradients for tool-use large language models, 2025.
\newblock URL \url{https://arxiv.org/abs/2509.21826}.

\bibitem[Liu et~al.(2025)Liu, Shen, Li, Ma, Jiang, Zhang, Fan, Li, Tang, and Luo]{liu2025surveytexttosqlerallms}
Xinyu Liu, Shuyu Shen, Boyan Li, Peixian Ma, Runzhi Jiang, Yuxin Zhang, Ju~Fan, Guoliang Li, Nan Tang, and Yuyu Luo.
\newblock A survey of text-to-sql in the era of llms: Where are we, and where are we going?, 2025.
\newblock URL \url{https://arxiv.org/abs/2408.05109}.

\bibitem[Lozhkov et~al.(2024)Lozhkov, Li, Allal, Cassano, Lamy-Poirier, Tazi, Tang, Pykhtar, Liu, Wei, Liu, Tian, Kocetkov, Zucker, Belkada, Wang, Liu, Abulkhanov, Paul, Li, Li, Risdal, Li, Zhu, Zhuo, Zheltonozhskii, Dade, Yu, Krauß, Jain, Su, He, Dey, Abati, Chai, Muennighoff, Tang, Oblokulov, Akiki, Marone, Mou, Mishra, Gu, Hui, Dao, Zebaze, Dehaene, Patry, Xu, McAuley, Hu, Scholak, Paquet, Robinson, Anderson, Chapados, Patwary, Tajbakhsh, Jernite, Ferrandis, Zhang, Hughes, Wolf, Guha, von Werra, and de~Vries]{lozhkov2024starcoder2stackv2}
Anton Lozhkov, Raymond Li, Loubna~Ben Allal, Federico Cassano, Joel Lamy-Poirier, Nouamane Tazi, Ao~Tang, Dmytro Pykhtar, Jiawei Liu, Yuxiang Wei, Tianyang Liu, Max Tian, Denis Kocetkov, Arthur Zucker, Younes Belkada, Zijian Wang, Qian Liu, Dmitry Abulkhanov, Indraneil Paul, Zhuang Li, Wen-Ding Li, Megan Risdal, Jia Li, Jian Zhu, Terry~Yue Zhuo, Evgenii Zheltonozhskii, Nii Osae~Osae Dade, Wenhao Yu, Lucas Krauß, Naman Jain, Yixuan Su, Xuanli He, Manan Dey, Edoardo Abati, Yekun Chai, Niklas Muennighoff, Xiangru Tang, Muhtasham Oblokulov, Christopher Akiki, Marc Marone, Chenghao Mou, Mayank Mishra, Alex Gu, Binyuan Hui, Tri Dao, Armel Zebaze, Olivier Dehaene, Nicolas Patry, Canwen Xu, Julian McAuley, Han Hu, Torsten Scholak, Sebastien Paquet, Jennifer Robinson, Carolyn~Jane Anderson, Nicolas Chapados, Mostofa Patwary, Nima Tajbakhsh, Yacine Jernite, Carlos~Muñoz Ferrandis, Lingming Zhang, Sean Hughes, Thomas Wolf, Arjun Guha, Leandro von Werra, and Harm de~Vries.
\newblock Starcoder 2 and the stack v2: The next generation, 2024.
\newblock URL \url{https://arxiv.org/abs/2402.19173}.

\bibitem[Lyu et~al.(2025)Lyu, Luo, Li, Ou, Sun, Qin, Shang, Song, and Zhu]{lyu2025sqlo1selfrewardheuristicdynamic}
Shuai Lyu, Haoran Luo, Ripeng Li, Zhonghong Ou, Jiangfeng Sun, Yang Qin, Xiaoran Shang, Meina Song, and Yifan Zhu.
\newblock Sql-o1: A self-reward heuristic dynamic search method for text-to-sql, 2025.
\newblock URL \url{https://arxiv.org/abs/2502.11741}.

\bibitem[Ma et~al.(2025)Ma, Zhuang, Xu, Jiang, Chen, and Guo]{ma2025sqlr1trainingnaturallanguage}
Peixian Ma, Xialie Zhuang, Chengjin Xu, Xuhui Jiang, Ran Chen, and Jian Guo.
\newblock Sql-r1: Training natural language to sql reasoning model by reinforcement learning, 2025.
\newblock URL \url{https://arxiv.org/abs/2504.08600}.

\bibitem[Mai et~al.(2025)Mai, Xu, Li, W, Wang, Hu, Zhang, and Zhang]{mai2025agentrlscalinglaw}
Xinji Mai, Haotian Xu, Zhong-Zhi Li, Xing W, Weinong Wang, Jian Hu, Yingying Zhang, and Wenqiang Zhang.
\newblock Agent rl scaling law: Agent rl with spontaneous code execution for mathematical problem solving, 2025.
\newblock URL \url{https://arxiv.org/abs/2505.07773}.

\bibitem[Mishra et~al.(2024)Mishra, Stallone, Zhang, Shen, Prasad, Soria, Merler, Selvam, Surendran, Singh, Sethi, Dang, Li, Wu, Zawad, Coleman, White, Lewis, Pavuluri, Koyfman, Lublinsky, de~Bayser, Abdelaziz, Basu, Agarwal, Zhou, Johnson, Goyal, Patel, Shah, Zerfos, Ludwig, Munawar, Crouse, Kapanipathi, Salaria, Calio, Wen, Seelam, Belgodere, Fonseca, Singhee, Desai, Cox, Puri, and Panda]{mishra2024granitecodemodelsfamily}
Mayank Mishra, Matt Stallone, Gaoyuan Zhang, Yikang Shen, Aditya Prasad, Adriana~Meza Soria, Michele Merler, Parameswaran Selvam, Saptha Surendran, Shivdeep Singh, Manish Sethi, Xuan-Hong Dang, Pengyuan Li, Kun-Lung Wu, Syed Zawad, Andrew Coleman, Matthew White, Mark Lewis, Raju Pavuluri, Yan Koyfman, Boris Lublinsky, Maximilien de~Bayser, Ibrahim Abdelaziz, Kinjal Basu, Mayank Agarwal, Yi~Zhou, Chris Johnson, Aanchal Goyal, Hima Patel, Yousaf Shah, Petros Zerfos, Heiko Ludwig, Asim Munawar, Maxwell Crouse, Pavan Kapanipathi, Shweta Salaria, Bob Calio, Sophia Wen, Seetharami Seelam, Brian Belgodere, Carlos Fonseca, Amith Singhee, Nirmit Desai, David~D. Cox, Ruchir Puri, and Rameswar Panda.
\newblock Granite code models: A family of open foundation models for code intelligence, 2024.
\newblock URL \url{https://arxiv.org/abs/2405.04324}.

\bibitem[Papicchio et~al.(2025)Papicchio, Rossi, Cagliero, and Papotti]{papicchio2025think2sqlreinforcellmreasoning}
Simone Papicchio, Simone Rossi, Luca Cagliero, and Paolo Papotti.
\newblock Think2sql: Reinforce llm reasoning capabilities for text2sql, 2025.
\newblock URL \url{https://arxiv.org/abs/2504.15077}.

\bibitem[Pourreza \& Rafiei(2023)Pourreza and Rafiei]{pourreza2023dinsqldecomposedincontextlearning}
Mohammadreza Pourreza and Davood Rafiei.
\newblock Din-sql: Decomposed in-context learning of text-to-sql with self-correction, 2023.
\newblock URL \url{https://arxiv.org/abs/2304.11015}.

\bibitem[Pourreza et~al.(2024)Pourreza, Li, Sun, Chung, Talaei, Kakkar, Gan, Saberi, Ozcan, and Arik]{pourreza2024chasesqlmultipathreasoningpreference}
Mohammadreza Pourreza, Hailong Li, Ruoxi Sun, Yeounoh Chung, Shayan Talaei, Gaurav~Tarlok Kakkar, Yu~Gan, Amin Saberi, Fatma Ozcan, and Sercan~O. Arik.
\newblock Chase-sql: Multi-path reasoning and preference optimized candidate selection in text-to-sql, 2024.
\newblock URL \url{https://arxiv.org/abs/2410.01943}.

\bibitem[Pourreza et~al.(2025)Pourreza, Talaei, Sun, Wan, Li, Mirhoseini, Saberi, and Arik]{pourreza2025reasoningsqlreinforcementlearningsql}
Mohammadreza Pourreza, Shayan Talaei, Ruoxi Sun, Xingchen Wan, Hailong Li, Azalia Mirhoseini, Amin Saberi, and Sercan~"O. Arik.
\newblock Reasoning-sql: Reinforcement learning with sql tailored partial rewards for reasoning-enhanced text-to-sql, 2025.
\newblock URL \url{https://arxiv.org/abs/2503.23157}.

\bibitem[Qin et~al.(2025)Qin, Chen, Fu, Chen, Peng, Hu, and Ye]{qin2025routerobustmultitasktuning}
Yang Qin, Chao Chen, Zhihang Fu, Ze~Chen, Dezhong Peng, Peng Hu, and Jieping Ye.
\newblock Route: Robust multitask tuning and collaboration for text-to-sql, 2025.
\newblock URL \url{https://arxiv.org/abs/2412.10138}.

\bibitem[Shang et~al.(2025)Shang, Liu, Zhu, Zhang, Xu, Guan, Zhang, Dong, Zhou, Zhang, Xin, Miao, Li, Yang, and Yang]{shang2025rstar2agentagenticreasoningtechnical}
Ning Shang, Yifei Liu, Yi~Zhu, Li~Lyna Zhang, Weijiang Xu, Xinyu Guan, Buze Zhang, Bingcheng Dong, Xudong Zhou, Bowen Zhang, Ying Xin, Ziming Miao, Scarlett Li, Fan Yang, and Mao Yang.
\newblock rstar2-agent: Agentic reasoning technical report, 2025.
\newblock URL \url{https://arxiv.org/abs/2508.20722}.

\bibitem[Team(2024)]{grattafiori2024llama3herdmodels}
Llama Team.
\newblock The llama 3 herd of models, 2024.
\newblock URL \url{https://arxiv.org/abs/2407.21783}.

\bibitem[team(2024)]{mistral_codestral}
Mistral~AI team.
\newblock Codestral | mistral ai, May 2024.
\newblock URL \url{https://mistral.ai/news/codestral}.

\bibitem[Wang et~al.(2025)Wang, Genadi, Bouardi, Wang, Koto, Liu, Baldwin, and Li]{wang2025agentflyextensiblescalablereinforcement}
Renxi Wang, Rifo~Ahmad Genadi, Bilal~El Bouardi, Yongxin Wang, Fajri Koto, Zhengzhong Liu, Timothy Baldwin, and Haonan Li.
\newblock Agentfly: Extensible and scalable reinforcement learning for lm agents, 2025.
\newblock URL \url{https://arxiv.org/abs/2507.14897}.

\bibitem[Xie et~al.(2024)Xie, Wu, and Zhou]{xie2024magsqlmultiagentgenerativeapproach}
Wenxuan Xie, Gaochen Wu, and Bowen Zhou.
\newblock Mag-sql: Multi-agent generative approach with soft schema linking and iterative sub-sql refinement for text-to-sql, 2024.
\newblock URL \url{https://arxiv.org/abs/2408.07930}.

\bibitem[Xue et~al.(2025)Xue, Zheng, Liu, Li, Zheng, Ma, and An]{xue2025simpletirendtoendreinforcementlearning}
Zhenghai Xue, Longtao Zheng, Qian Liu, Yingru Li, Xiaosen Zheng, Zejun Ma, and Bo~An.
\newblock Simpletir: End-to-end reinforcement learning for multi-turn tool-integrated reasoning, 2025.
\newblock URL \url{https://arxiv.org/abs/2509.02479}.

\bibitem[Yang et~al.(2024)Yang, Hui, Yang, Yang, Lin, and Zhou]{yang2024synthesizingtexttosqldataweak}
Jiaxi Yang, Binyuan Hui, Min Yang, Jian Yang, Junyang Lin, and Chang Zhou.
\newblock Synthesizing text-to-sql data from weak and strong llms, 2024.
\newblock URL \url{https://arxiv.org/abs/2408.03256}.

\bibitem[Yao et~al.(2025)Yao, Sun, Borchmann, Shen, Deng, Zhai, Zhang, Li, and He]{yao2025arctictext2sqlr1simplerewardsstrong}
Zhewei Yao, Guoheng Sun, Lukasz Borchmann, Zheyu Shen, Minghang Deng, Bohan Zhai, Hao Zhang, Ang Li, and Yuxiong He.
\newblock Arctic-text2sql-r1: Simple rewards, strong reasoning in text-to-sql, 2025.
\newblock URL \url{https://arxiv.org/abs/2505.20315}.

\bibitem[Yu et~al.(2025)Yu, Zhang, Zhu, Yuan, Zuo, Yue, Dai, Fan, Liu, Liu, Liu, Lin, Lin, Ma, Sheng, Tong, Zhang, Zhang, Zhang, Zhu, Zhu, Chen, Chen, Wang, Yu, Song, Wei, Zhou, Liu, Ma, Zhang, Yan, Qiao, Wu, and Wang]{yu2025dapoopensourcellmreinforcement}
Qiying Yu, Zheng Zhang, Ruofei Zhu, Yufeng Yuan, Xiaochen Zuo, Yu~Yue, Weinan Dai, Tiantian Fan, Gaohong Liu, Lingjun Liu, Xin Liu, Haibin Lin, Zhiqi Lin, Bole Ma, Guangming Sheng, Yuxuan Tong, Chi Zhang, Mofan Zhang, Wang Zhang, Hang Zhu, Jinhua Zhu, Jiaze Chen, Jiangjie Chen, Chengyi Wang, Hongli Yu, Yuxuan Song, Xiangpeng Wei, Hao Zhou, Jingjing Liu, Wei-Ying Ma, Ya-Qin Zhang, Lin Yan, Mu~Qiao, Yonghui Wu, and Mingxuan Wang.
\newblock Dapo: An open-source llm reinforcement learning system at scale, 2025.
\newblock URL \url{https://arxiv.org/abs/2503.14476}.

\bibitem[Yu et~al.(2019)Yu, Zhang, Yang, Yasunaga, Wang, Li, Ma, Li, Yao, Roman, Zhang, and Radev]{yu2019SPIDERlargescalehumanlabeleddataset}
Tao Yu, Rui Zhang, Kai Yang, Michihiro Yasunaga, Dongxu Wang, Zifan Li, James Ma, Irene Li, Qingning Yao, Shanelle Roman, Zilin Zhang, and Dragomir Radev.
\newblock Spider: A large-scale human-labeled dataset for complex and cross-domain semantic parsing and text-to-sql task, 2019.
\newblock URL \url{https://arxiv.org/abs/1809.08887}.

\bibitem[Yue et~al.(2025)Yue, Dong, Gao, He, Chai, Yin, and Lin]{yue2025promotingefficientreasoningverifiable}
Chuhuai Yue, Chengqi Dong, Yinan Gao, Hang He, Jiajun Chai, Guojun Yin, and Wei Lin.
\newblock Promoting efficient reasoning with verifiable stepwise reward, 2025.
\newblock URL \url{https://arxiv.org/abs/2508.10293}.

\bibitem[Zeng et~al.(2025)Zeng, Wei, Brown, Frunza, Nevmyvaka, and Hong]{zeng2025reinforcingmultiturnreasoningllm}
Siliang Zeng, Quan Wei, William Brown, Oana Frunza, Yuriy Nevmyvaka, and Mingyi Hong.
\newblock Reinforcing multi-turn reasoning in llm agents via turn-level credit assignment, 2025.
\newblock URL \url{https://arxiv.org/abs/2505.11821}.

\bibitem[Zhai et~al.(2025)Zhai, Xu, He, and Yao]{zhai2025excotoptimizingreasoningtexttosql}
Bohan Zhai, Canwen Xu, Yuxiong He, and Zhewei Yao.
\newblock Excot: Optimizing reasoning for text-to-sql with execution feedback, 2025.
\newblock URL \url{https://arxiv.org/abs/2503.19988}.

\bibitem[Zhang et~al.(2025)Zhang, Geng, Yu, Yin, Zhang, Tan, Zhou, Li, Xue, Li, Zhou, Chen, Zhang, Fan, Wang, Huang, Liao, Wang, Yang, Ji, Littman, Wang, Yan, Torr, and Bai]{zhang2025landscapeagenticreinforcementlearning}
Guibin Zhang, Hejia Geng, Xiaohang Yu, Zhenfei Yin, Zaibin Zhang, Zelin Tan, Heng Zhou, Zhongzhi Li, Xiangyuan Xue, Yijiang Li, Yifan Zhou, Yang Chen, Chen Zhang, Yutao Fan, Zihu Wang, Songtao Huang, Yue Liao, Hongru Wang, Mengyue Yang, Heng Ji, Michael Littman, Jun Wang, Shuicheng Yan, Philip Torr, and Lei Bai.
\newblock The landscape of agentic reinforcement learning for llms: A survey, 2025.
\newblock URL \url{https://arxiv.org/abs/2509.02547}.

\bibitem[Zhao et~al.(2025)Zhao, Wang, Ma, Kong, Yang, Tuo, Shi, Zhai, and Cai]{zhao2025muarlmultiturnuserinteractingagent}
Weikang Zhao, Xili Wang, Chengdi Ma, Lingbin Kong, Zhaohua Yang, Mingxiang Tuo, Xiaowei Shi, Yitao Zhai, and Xunliang Cai.
\newblock Mua-rl: Multi-turn user-interacting agent reinforcement learning for agentic tool use, 2025.
\newblock URL \url{https://arxiv.org/abs/2508.18669}.

\end{thebibliography}
\bibliographystyle{iclr2026_conference}

\newpage

\appendix

\section{Use of LLMs}
In this work, we use Large Language Models (LLMs) for text refinement and grammar checking. LLMs help improve the clarity, coherence, and grammatical accuracy of the generated content, ensuring the final text meets academic standards. Their use is limited to enhancing written content, without influencing the research methodology or decision-making process.
\section{LLM Response Rollout with Multi-Turn Sql Excursions Calls}\label{app:algr1}

The algorithm describes the response generation process of a generative model (e.g., LLM) based on multi-turn interactions. The core idea of the algorithm is to progressively generate a response sequence based on the user's input and previous responses. In each generation step, the model evaluates the current output and interacts with external tools for validation (e.g., executing SQL queries). The results returned by the tool are then integrated into the generated response. The entire process is conducted within a maximum action budget to ensure that the final output meets the problem's requirements and is validated for accuracy. After each round, the model adjusts its output based on the results, continuing until a complete response is achieved or the budget limit is reached.

\begin{algorithm}[htbp]
\caption{LLM Response Rollout with Multi-Turn SQL Execution Tool Calls}
\label{alg:llm-response-rollout}
\textbf{Require:} Input query $x$, policy model $\pi_\theta$, SQL execution tool $\mathcal{T}$, maximum action budget $B$. \\
\textbf{Ensure:} Final response $y$.

\begin{algorithmic}[1]
\State Initialize rollout sequence $y \gets \emptyset$
\State Initialize action count $b \gets 0$
\While{$b < B$}
    \State Initialize current action LLM rollout sequence $y_b \gets \emptyset$
    \While{True}
        \State Generate response token $y_t \sim \pi_\theta(\cdot | x, y, y_b)$
        \State Append $y_t$ to rollout sequence $y_b \gets y_b + y_t$
        \If{$y_t \in \langle \text{tool call} \rangle, \langle \text{tool response} \rangle, \langle \text{eos} \rangle$}
            \State \textbf{break}
        \EndIf
    \EndWhile
    \State $y \gets y + y_b$
    \If{$\langle \text{tool call} \rangle$ detected in $y_b$}
        \State Extract SQL query $q \gets \text{Parse}(y_b, \langle \text{tool call} \rangle, \langle \text{tool call} \rangle)$
        \State Retrieve SQL query results $d \gets \mathcal{T}(q)$
        \State Insert $d$ into rollout $y \gets y + \langle \text{tool response} \rangle d \langle \text{tool response} \rangle$
    \ElsIf{$\langle \text{tool response} \rangle$ detected in $y_b$}
        \State \textbf{return} final generated response $y$
    \Else
        \State Ask for rethink $y \gets y + \langle \text{My action is not correct. Let me rethink.} \rangle$
    \EndIf
    \State Increment action count $b \gets b + 1$
\EndWhile
\State \textbf{return} final generated response $y$
\end{algorithmic}
\end{algorithm}

\newpage
\section{Prompt and Case Study}
Prompts used while training and several cases are presented as follows.
\subsection{System Prompt}\label{system}
\begin{tcolorbox}[
    colback=cyan!5,
    colframe=cyan!75,
    width=\textwidth,
    arc=1mm,
    boxrule=0.5pt,
    title=\textbf{System Prompt of Tool },
    halign title=center,
    breakable
]

\textbf{\#\#Tools} \\

You may call one or more functions to assist with the user query. \\

You are provided with function signatures within \texttt{<tools></tools>} XML tags:
    
    
    

\begin{lstlisting}[
    breaklines=true,
    linewidth=0.9\textwidth,
    basicstyle=\footnotesize\ttfamily,
    xleftmargin=1em,
    xrightmargin=1em
]
<tools>
  {"name": "sql-execute_sql_query", "description": "Execute SQL query and return partial results containing column names (maximum 10 records).
    
    Args:db_name (str): The name of the database.
      sql (str): The SQL query to execute.
    
    Returns:Dict[str, Union[List[Dict], Dict, None]]: A dictionary containing 'columns' and 'data' of the query (maximum of 10 records).
    
    Raises: TimeoutError: If the query execution exceeds the timeout.
      sqlite3.Error: If an error occurs during the query execution.
    ", 
    "parameters": {
      "type": "object", 
      "properties": {
        "db_name": {"title": "Db Name", "type": "string"}, 
        "sql": {"title": "Sql", "type": "string"}
      }, 
      "required": ["db_name", "sql"]
    }
}
</tools>
\end{lstlisting}

For each function call, return a JSON object with function name and arguments within \texttt{<tool\_call></tool\_call>} XML tags:

\begin{lstlisting}[
    breaklines=true,
    linewidth=0.9\textwidth,
    basicstyle=\footnotesize\ttfamily,
    xleftmargin=1em,
    xrightmargin=1em
]
<tool_call>
  {"name": <function-name>, "arguments": <args-json-object>}
</tool_call}

\end{lstlisting}

\end{tcolorbox}

\newpage
\subsection{User Prompt}

\begin{tcolorbox}[
    colback=cyan!5,
    colframe=cyan!75,
    width=\textwidth,
    arc=1mm,
    boxrule=0.5pt,
    title=\textbf{Prompts used while training and evaluating},
    halign title=center,
    breakable
]

You are a helpful SQL expert assistant. You should first think about how to write the SQL query by analyzing the question, database schema, and external knowledge, then validate your SQL with the tool until it is correct. Finally, you provide the final SQL query in \texttt{<answer>} \texttt{</answer>}.

\textbf{Task Configuration} \\
\textbf{Database Engine:} SQLite \\
\textbf{Database:} \{db\_id\} \\
\textbf{Database Schema:} \{schema\} \\
\textbf{User Question:} \{external\_knowledge\} + \{question\}? 

\textbf{Requirements}
\begin{enumerate}
    \item \textbf{Precision:} Make sure you only output the information that is asked in the question. If the question asks for a specific column, make sure to only include that column in the SELECT clause, nothing more.
    \item \textbf{Completeness:} The generated query should return all of the information asked in the question without any missing or extra information.
    \item \textbf{Correctness:} Before generating the final SQL query, please think through the steps of how to write the query. Validate your SQL through tool testing.
\end{enumerate}

\textbf{Output Format:} \\
\textbf{Important:} Use EITHER thinking + tool calls OR thinking + final answer. Do not mix the structures.

\textbf{Option A (when validation needed):} 

\begin{lstlisting}[
    breaklines=true,
    linewidth=0.9\textwidth,
    basicstyle=\footnotesize\ttfamily,
    xleftmargin=1em,
    xrightmargin=1em
]
<think> Your analysis... </think>
[Tool calls for validation]

\end{lstlisting}

\textbf{Option B (final answer):} 


\begin{lstlisting}[
    breaklines=true,
    linewidth=0.9\textwidth,
    basicstyle=\footnotesize\ttfamily,
    xleftmargin=1em,
    xrightmargin=1em
]
<think> Your final analysis... </think>
<answer> 
```sql
YOUR_SQL_QUERY
</answer>

\end{lstlisting}

\end{tcolorbox}
\newpage
\subsection{Case study}

\begin{tcolorbox}[
    colback=cyan!5,
    colframe=cyan!75,
    width=\textwidth,
    arc=1mm,
    boxrule=0.5pt,
    title=\textbf{Case1 Single Turn},
    halign title=center,
    breakable
]
 \section*{User Question}
    Exclusively virtual refers to Virtual = 'F'. How many schools with an average score in Math greater than 400 in the SAT test are exclusively virtual?

    \section*{Model Response}

\textless think\textgreater 

Okay, let's tackle this problem. The user is asking for the number of schools that are exclusively virtual (Virtual = 'F') and have an average math score greater than 400 in the SAT test.... So the query should be correct.

\textless /think\textgreater

\textless tool\_call\textgreater

\begin{lstlisting}
{
  "name": "sql-execute_sql_query",
  "arguments": {
    "db_name": "california_schools",
    "sql": "SELECT COUNT(*) FROM satscores JOIN schools ON satscores.cds = schools.CDSCode WHERE schools.Virtual = 'F' AND satscores.AvgScrMath > 400;"
  }
}
\end{lstlisting}
\textless /tool\_call\textgreater

\textless tool\_response\textgreater
\begin{lstlisting}
The result is:{
  "columns": [
    "COUNT(*)"
  ],
  "data": [
    {
      "COUNT(*)": 4
    }
  ]
}
\end{lstlisting}
\textless /tool\_response\textgreater

\textless think\textgreater 

Okay, let me check the user's question again. They want the number of schools that are exclusively virtual (Virtual = 'F') and have an average math score greater than 400 in the SAT.....The result from the tool is 4, so that's the answer.

\textless /think\textgreater

\textless answer\textgreater

```sql
\begin{lstlisting}[language=SQL]
SELECT COUNT(*) FROM satscores JOIN schools ON satscores.cds = schools.CDSCode WHERE schools.Virtual = 'F' AND satscores.AvgScrMath > 400; \end{lstlisting}

```

    \textless /answer\textgreater

\end{tcolorbox}

\begin{tcolorbox}[
    colback=cyan!5,
    colframe=cyan!75,
    width=\textwidth,
    arc=1mm,
    boxrule=0.5pt,
    title=\textbf{Case2 Multi-Turn},
    halign title=center,
    breakable
]
 \section*{User Question} calcium refers to element = 'ca'; label = '+' mean molecules are carcinogenic; label = '-' means molecules are non-carcinogenic; MAX(label).Among the molecules with element Calcium, are they mostly carcinogenic or non carcinogenic?

\section*{Model Response}

\textbf{Turn 1}

\textless think\textgreater 

Alright, let's solve this problem. The user wants to know whether molecules containing calcium element (element 'Ca') are mostly carcinogenic or not. Label '+' means carcinogenic, '-' means non-carcinogenic. So, we need to find the maximum label value for molecules containing calcium... So the final SQL query will be to select the maximum label from molecules that contain calcium atoms.

\textless /think\textgreater

\textless tool\_call\textgreater

\begin{lstlisting}
{
  "name": "sql-execute_sql_query",
  "arguments": {
    "db_name": "toxicology",
    "sql": "SELECT MAX(m.label) AS max_label FROM molecule m JOIN atom a ON m.molecule_id = a.molecule_id WHERE a.element = 'Ca';"
  }
}
\end{lstlisting}
\textless /tool\_call\textgreater

\textless tool\_response\textgreater
\begin{lstlisting}
The result is:: {
  "columns": [
    "max_label"
  ],
  "data": [
    {
      "max_label": null
    }
  ]
}
\end{lstlisting}
\textless /tool\_response\textgreater

\textbf{Turn 2}

\textless think\textgreater 

Hmm, the SQL query result returns max\_label as null. This is unexpected. Let me check the query again. Maybe the label in the molecule table is stored as a string, but perhaps the query didn't correctly select the label... Let me try the corrected query.

\textless /think\textgreater

\textless tool\_call\textgreater

\begin{lstlisting}
{
  "name": "sql-execute_sql_query",
  "arguments": {
    "db_name": "toxicology",
    "sql": "SELECT MAX(m.label) AS max_label FROM molecule m JOIN atom a ON m.molecule_id = a.molecule_id WHERE a.element = 'CA';"
  }
}
\end{lstlisting}
\textless /tool\_call\textgreater

\textless tool\_response\textgreater
\begin{lstlisting}
The result is: {
  "columns": [
    "max_label"
  ],
  "data": [
    {
      "max_label": null
    }
  ]
}
\end{lstlisting}
\textless /tool\_response\textgreater

\textbf{Turn 3}

\textless think\textgreater 

Wait, the result is still null. Let me think again. Maybe the label is stored as a string in the molecule table, but the query didn't select it correctly. Let me try the corrected query.

\textless /think\textgreater

\textless tool\_call\textgreater

\begin{lstlisting}
{
  "name": "sql-execute_sql_query",
  "arguments": {
    "db_name": "toxicology",
    "sql": "SELECT MAX(m.label) AS max_label FROM molecule m JOIN atom a ON m.molecule_id = a.molecule_id WHERE a.element = 'ca';"
  }
}
\end{lstlisting}
\textless /tool\_call\textgreater

\textless tool\_response\textgreater
\begin{lstlisting}
The result is: {
  "columns": [
    "max_label"
  ],
  "data": [
    {
      "max_label": "-"
    }
  ]
}
\end{lstlisting}
\textless /tool\_response\textgreater

\textbf{Turn 4}

\textless think\textgreater 

Okay, so after adjusting the query to use `ca' as the element, the result is that the max\_label is `-'. That means among the molecules with calcium atoms, the maximum label is `-', which indicates they are non-carcinogenic. Therefore, the answer should be that they are mostly non-carcinogenic.

\textless /think\textgreater

\textless answer\textgreater

```sql
\begin{lstlisting}[language=SQL]
SELECT MAX(m.label) AS max_label FROM molecule m JOIN atom a ON m.molecule_id = a.molecule_id WHERE a.element = 'ca'; \end{lstlisting}
```

    \textless /answer\textgreater

\end{tcolorbox}

\begin{tcolorbox}[
    colback=cyan!5,
    colframe=cyan!75,
    width=\textwidth,
    arc=1mm,
    boxrule=0.5pt,
    title=\textbf{Case3 Multi-Turn},
    halign title=center,
    breakable
]
  \section*{User Question} What is the number of SAT test takers of the schools with the highest FRPM count for K-12 students?

\section*{Model Response}

\textbf{Turn 1}

\textless think\textgreater 

Alright, let me solve this problem. The user is asking about the number of SAT test takers at schools with the highest FRPM count for K-12 students.
First, I need to determine which tables are involved....
Let me verify using the tool.

\textless /think\textgreater

\textless tool\_call\textgreater

\begin{lstlisting}
{
  "name": "sql-execute_sql_query",
  "arguments": {
    "db_name": "california_schools",
    "sql": "SELECT SUM(s.NumTstTakr) FROM satscores s JOIN frpm f ON s.cds = f.CDSCode WHERE f.\"FRPM Count (K-12)\" = (SELECT MAX(\"FRPM Count (K-12)\") FROM frpm);"
  }
}
\end{lstlisting}
\textless /tool\_call\textgreater

\textless tool\_response\textgreater
\begin{lstlisting}
{
  "columns": [
    "SUM(s.NumTstTakr)"
  ],
  "data": [
    {
      "SUM(s.NumTstTakr)": null
    }
  ]
}
\end{lstlisting}
\textless /tool\_response\textgreater

\textbf{Turn 2}

\textless think\textgreater 

Hmm, the SQL query returned a sum result of null. This is not right. Let me check the query again. Maybe there's an error in the column names or table names...
Let me test again.

\textless /think\textgreater

\textless tool\_call\textgreater

\begin{lstlisting}
{
  "name": "sql-execute_sql_query",
  "arguments": {
    "db_name": "california_schools",
    "sql": "SELECT SUM(s.NumTstTakr) FROM satscores s JOIN frpm f ON s.cds = f.CDSCode WHERE f.\"FRPM Count (K-12)\" = (SELECT MAX(\"FRPM Count (K-12)\") FROM frpm);"
  }
}
\end{lstlisting}
\textless /tool\_call\textgreater

\textless tool\_response\textgreater
\begin{lstlisting}
{
  "columns": [
    "SUM(s.NumTstTakr)"
  ],
  "data": [
    {
      "SUM(s.NumTstTakr)": null
    }
  ]
}
\end{lstlisting}
\textless /tool\_response\textgreater

\textbf{Turn 3}

\textless think\textgreater 

Hmm, the result is still null. This is strange. Maybe there's an issue with the query structure. I need to check if the column name in the frpm table is really named "FRPM Count (K-12)" or if there's a spelling error.
Let me try again.

\textless /think\textgreater

\textless tool\_call\textgreater

\begin{lstlisting}
{
  "name": "sql-execute_sql_query",
  "arguments": {
    "db_name": "california_schools",
    "sql": "SELECT SUM(s.NumTstTakr) FROM satscores s JOIN frpm f ON s.cds = f.CDSCode WHERE f.\"FRPM Count (K-12)\" = (SELECT MAX(\"FRPM Count (K-12)\") FROM frpm);"
  }
}
\end{lstlisting}
\textless /tool\_call\textgreater

\textless tool\_response\textgreater
\begin{lstlisting}
{
  "columns": [
    "SUM(s.NumTstTakr)"
  ],
  "data": [
    {
      "SUM(s.NumTstTakr)": null
    }
  ]
}
\end{lstlisting}
\textless /tool\_response\textgreater

\textbf{Turn 4}

\textless think\textgreater

Maybe I should first check what the maximum value of FRPM Count (K-12) in the frpm table is. Let me write a query to get that value.

\textless /think\textgreater

\textless tool\_call\textgreater

\begin{lstlisting}
{
  "name": "sql-execute_sql_query",
  "arguments": {
    "db_name": "california_schools",
    "sql": "SELECT SUM(s.NumTstTakr) FROM satscores s JOIN frpm f ON s.cds = f.CDSCode WHERE f.\"FRPM Count (K-12)\" = (SELECT MAX(\"FRPM Count (K-12)\") FROM frpm);"
  }
}
\end{lstlisting}
\textless /tool\_call\textgreater

\textless tool\_response\textgreater
\begin{lstlisting}
{
  "columns": [
    "SUM(s.NumTstTakr)"
  ],
  "data": [
    {
      "SUM(s.NumTstTakr)": null
    }
  ]
}
\end{lstlisting}
\textless /tool\_response\textgreater

\textless think\textgreater 

Maybe there's an issue with the column names or the tables not joining correctly. Let me think again. Maybe I should check if the satscores table has data for schools in the frpm table.... I think this should work. Let me use this as the final query.

\textless /think\textgreater

\textless tool\_call\textgreater

\begin{lstlisting}
{
  "name": "sql-execute_sql_query",
  "arguments": {
    "db_name": "california_schools",
    "sql": "SELECT SUM(s.NumTstTakr) FROM satscores s JOIN (SELECT CDSCode FROM frpm WHERE \"FRPM Count (K-12)\" = (SELECT MAX(\"FRPM Count (K-12)\") FROM frpm)) AS top_frpm ON s.cds = top_frpm.CDSCode;"
  }
}
\end{lstlisting}
\textless /tool\_call\textgreater

\textless tool\_response\textgreater
\begin{lstlisting}
{
  "columns": [
    "SUM(s.NumTstTakr)"
  ],
  "data": [
    {
      "SUM(s.NumTstTakr)": 217547
    }
  ]
}
\end{lstlisting}
\textless /tool\_response\textgreater

\textbf{Turn 5}

\textless think\textgreater 

The tool response shows a result of 217547, which seems to be a valid number.

\textless /think\textgreater

\textless answer\textgreater

```sql
\begin{lstlisting}[language=SQL]
SELECT SUM(s.NumTstTakr) FROM satscores s JOIN (SELECT CDSCode FROM frpm WHERE "FRPM Count (K-12)" = (SELECT MAX("FRPM Count (K-12)") FROM frpm)) AS top_frpm ON s.cds = top_frpm.CDSCode; \end{lstlisting}
    ```
    
\textless /answer\textgreater

\end{tcolorbox}

\end{document}